\documentclass[10pt,letterpaper]{article}
\usepackage{cogsci}
\cogscifinalcopy
\usepackage[utf8]{inputenc} 
\usepackage[T1]{fontenc}    
\usepackage{microtype}
\usepackage{times,latexsym}
\usepackage{graphicx} \graphicspath{{figures/}}
\usepackage{amsmath,nicefrac}
\usepackage{acronym}
\usepackage[breaklinks,colorlinks,citecolor=black,bookmarks=true]{hyperref}
\usepackage{balance}
\usepackage{xspace}
\usepackage{setspace}
\usepackage[skip=3pt,font=small]{subcaption}
\usepackage[skip=3pt,font=small]{caption}
\usepackage[dvipsnames,svgnames,x11names,table]{xcolor}
\usepackage[capitalise,noabbrev,nameinlink]{cleveref}
\usepackage{booktabs,tabularx,colortbl,multirow,multicol,array,makecell,tabularray}
\usepackage{enumitem}
\usepackage[misc]{ifsym}
\usepackage{dblfloatfix}

\makeatletter
\DeclareRobustCommand\onedot{\futurelet\@let@token\@onedot}
\def\@onedot{\ifx\@let@token.\else.\null\fi\xspace}
\def\eg{\emph{e.g}\onedot}

\def\vs{\emph{vs}\onedot}

\makeatother

\makeatletter
\renewcommand{\paragraph}{%
    \@startsection{paragraph}{4}%
    {\z@}{0ex \@plus 0ex \@minus 0ex}{-1em}%
    {\normalfont\normalsize\bfseries}%
}
\makeatother

\usepackage[backend=biber,style=apa,natbib=true,annotation=false,backref=true]{biblatex}
\crefname{algorithm}{Alg.}{Algs.}
\Crefname{algocf}{Algorithm}{Algorithms}
\crefname{section}{Sec.}{Secs.}
\Crefname{section}{Section}{Sections}
\crefname{table}{Tab.}{Tabs.}
\Crefname{table}{Table}{Tables}
\crefname{figure}{Fig.}{Figs.}
\Crefname{figure}{Figure}{Figures}
\crefname{equation}{Eq.}{Eqs.}
\Crefname{equation}{Equation}{Equations}
\crefname{appendix}{Appx.}{Appxs.}
\Crefname{appendix}{Appendix}{Appendices}

\AtBeginDocument{
    \acrodef{vlm}[VLM]{Vision Language Model}
    \acrodef{llm}[LLM]{Large Language Model}
    \acrodef{ai}[AI]{Artificial Intelligence}
    \acrodef{cc}[CC]{Common Cause}
    \acrodef{ce}[CE]{Common Effect}
    \acrodef{rl}[RL]{Reinforcement Learning}
}

\title{Grounding Before Generalizing: How AI Differs from Humans in Causal Transfer}

\author{%
    Liangru Xiang$^{1,4,\,*}$, Yuxi Ma$^{2,3,4,5\,*}$, Zhihao Cao$^{1,4\,*}$, Yixin Zhu$^{3,2,4,5\,\textrm{\Letter}}$, and Song-Chun Zhu$^{1,2,4\,\textrm{\Letter}}$
    \vspace{6pt}\\\normalfont
    \small $^\star{}$Equal contributors\quad
    \small Project Website: \url{https://causal-openlock.github.io}\\
    \small $^1$ Department of Automation, Tsinghua University\quad{}
    \small $^2$ Institute for Artificial Intelligence, Peking University\\
    \small $^3$ School of Psychological and Cognitive Sciences, Peking University\quad
    \small $^4$ State Key Laboratory of General Artificial Intelligence\\
    \small $^5$ Beijing Key Laboratory of Behavior and Mental Health, Peking University\quad{}
    \vspace{-6pt}
}

\begin{document}

\maketitle

\begin{abstract}
Extracting abstract causal structures and applying them to novel situations is a hallmark of human intelligence \citep{griffiths2005structure,holyoak2011causal,lake2017building}.
While \acp{llm} and \acp{vlm} have shown strong performance on a wide range of reasoning tasks \citep{brown2020language,xu2025towards}, their capacity for interactive causal learning---inducing latent structures through sequential exploration and transferring them across contexts---remains uncharacterized. Human learners accomplish such transfer after minimal exposure, whereas classical \ac{rl} agents fail catastrophically \citep{edmonds2018human}.
Whether state-of-the-art \ac{ai} models possess human-like mechanisms for abstract causal structure transfer is an open question.
Using the OpenLock paradigm \citep{edmonds2018human} requiring sequential discovery of \ac{cc} and \ac{ce} structures, here we show that models exhibit fundamentally delayed or absent transfer: even successful models require initial environmental-specific mapping---what we term environmental grounding---before efficiency gains emerge, whereas humans leverage prior structural knowledge from the very first solution attempt.
In the text-only condition, models matched or exceeded human discovery efficiency. In contrast, visual information---in both the image-only and text-and-image conditions---overall degraded rather than enhanced performance, revealing a broad reliance on symbolic processing rather than integrated multimodal reasoning. Models further exhibited systematic \ac{cc}/\ac{ce} asymmetries absent in humans, suggesting heuristic biases rather than direction-neutral causal abstraction.
These findings reveal that large-scale statistical learning does not produce the decontextualized causal schemas underpinning human analogical reasoning, establishing grounding-dependent transfer as a fundamental limitation of current \acp{llm} and \acp{vlm}.

\end{abstract}

\section{Introduction}

\acfp{llm} and \acfp{vlm} have achieved remarkable success across tasks ranging from natural language understanding to visual reasoning and mathematical problem-solving \citep{brown2020language,achiam2023gpt,openai2023gpt4v,anthropic2024claude,team2023gemini,liu2025deepseek,zhang2026proposing}. Yet a fundamental question remains: do these models engage in the kind of active causal learning and structural abstraction that characterizes human intelligence? While humans readily discover causal relationships through interaction and transfer this knowledge to new contexts, whether state-of-the-art \ac{ai} models possess comparable capabilities remains largely unexplored.

Causal structure transfer---the ability to recognize and apply abstract relational patterns across different domains---represents a cornerstone of human cognition \citep{holyoak1996mental,holyoak2010analogical,holyoak2011causal,griffiths2005structure,griffiths2009theory,lu2008bayesian}. Consider a smartphone unlockable via fingerprint, facial recognition, or passcode: this exemplifies a \textbf{many-to-one \acf{ce} structure}, where multiple independent causes converge on a single effect. Once grasped, the principle generalizes---a learner intuitively expects other secure systems to offer ``multiple pathways to authorization.'' Conversely, a power strip failure that simultaneously cuts power to a lamp, laptop, and television instantiates a \textbf{one-to-many \acf{cc} structure}, where a single cause propagates to multiple effects. Recognizing such patterns guides efficient learning: one seeks a central breaker rather than inspecting each device individually. Crucially, such structural abstraction allows agents to navigate unfamiliar environments without relearning from scratch.

Empirical evidence confirms that humans excel at causal structure transfer in interactive settings, exhibiting marked efficiency gains when transitioning between structurally similar environments \citep{edmonds2018human,edmonds2019decomposing,edmonds2020theory}. Remarkably, this transfer occurs after minimal exposure---often a single episode. By contrast, traditional \ac{rl} agents fail catastrophically on the same tasks despite orders of magnitude more training data, demonstrating that purely associative mechanisms cannot capture genuine structural abstraction.

Whether \acp{vlm} and \acp{llm} can bridge this gap is a pressing open question. These models are trained on vast corpora encoding rich causal and relational knowledge: \acp{vlm} integrate visual perception with language understanding, potentially enabling causal pattern extraction from visual scenes \citep{openai2023gpt4v,anthropic2024claude,team2023gemini}, while \acp{llm} have shown promise in linguistic causal reasoning \citep{kiciman2023causal,jin2024can,jin2023cladder} and structural pattern extraction via in-context learning \citep{brown2020language}. Yet these capacities have not been tested in interactive settings that demand active exploration and discovery of latent causal structure through sequential decision-making---precisely the conditions under which human transfer is most striking.

\begin{figure}[t!]
    \centering
    \includegraphics[width=\linewidth]{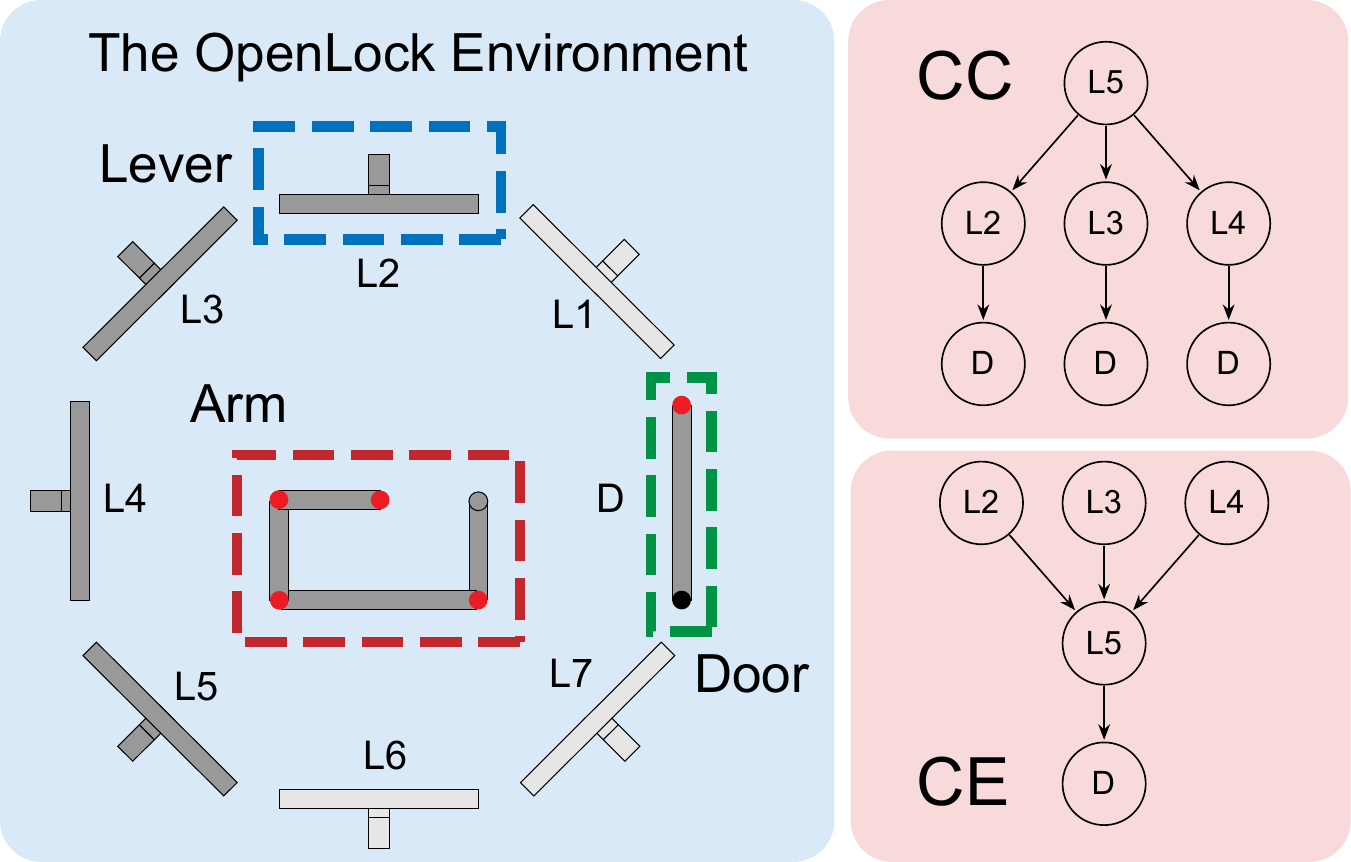}
    \caption{\textbf{OpenLock environment and causal structure schematics.} The virtual environment (\textit{left}) contains seven levers and one door; agents discover sequential causal dependencies through active exploration, with lever positions, colors, and labels varying across environments while the underlying causal topology is preserved. The latent causal graphs (\textit{right}) differ across the two experimental conditions: \acs{cc} instantiates a \textit{divergent} one-to-many structure in which a single first-stage lever ($L_5$) enables multiple independent second-stage paths to the door, whereas \acs{ce} instantiates a \textit{convergent} many-to-one structure in which multiple first-stage levers funnel through a shared bottleneck lever ($L_5$) before reaching the door.}
    \label{fig:env}
\end{figure}

We address this gap by adapting the OpenLock paradigm \citep{edmonds2018human}---originally developed to benchmark human causal transfer against \ac{rl} agents---to systematically probe four state-of-the-art \ac{ai} models (GPT-5.2, Claude-4.5-Sonnet, Gemini-3-Flash, and DeepSeek-V3.2). The layout of the environment is shown in \cref{fig:env}. This framework affords a systematic dissociation between local causal discovery---finding solutions within a single environment---and genuine structural abstraction---transferring learned relational schemas to perceptually novel environments. By directly comparing model behavior against the human data reported in \citet{edmonds2018human}, we evaluate models across three dimensions: (i) efficiency in causal discovery, (ii) the influence of input modality on active reasoning, and (iii) the transfer of learned structures to new environments.

Our findings reveal a fundamental divergence between human and model behavior. In the text-only condition, models matched or exceeded human discovery efficiency within a single environment. However, visual information---in both the image-only and text-and-image conditions---degraded rather than enhanced model performance, suggesting that current \acp{vlm} rely on symbolic processing rather than integrated visual-language reasoning. Most critically, while humans immediately exploited prior structural knowledge upon entering a new environment \citep{edmonds2018human}---exhibiting strong positive transfer from the very first solution attempt---no model showed such \textit{a priori} benefit. Instead, efficiency gains emerged only after models independently discovered an initial solution in the new context, a pattern consistent with post-hoc environmental grounding rather than genuine structural abstraction. We use ``grounding' here strictly in the sense of environmental mapping between abstract structure and situational tokens, distinct from multimodal grounding. Models further exhibited systematic performance asymmetries between \ac{cc} and \ac{ce} configurations that were absent in human learners, suggesting reliance on heuristic biases encoded during training rather than on direction-neutral causal representations.

\section{The OpenLock Paradigm}

The OpenLock task \citep{edmonds2018human} provides a controlled environment for studying interactive causal induction and structure transfer. Originally designed to compare human learners against \ac{rl} agents---revealing that humans transfer causal structures after minimal exposure while \ac{rl} agents fail to do so even with extensive training---the paradigm offers an ideal testbed for probing whether modern \ac{ai} models exhibit similarly human-like abstraction. Each environment contains eight interactive components: seven levers and one door. Success requires discovering three unique solutions within a budget of 30 attempts, where each attempt is strictly limited to a three-action sequence: two lever manipulations followed by a door-opening attempt. A model is considered successful if it identifies all three solutions within this budget. Crucially, the underlying causal graph is latent---agents must infer it through active exploration rather than passive observation.

The two experimental variants instantiate distinct causal graph topologies over four active components (three levers and the door); the remaining four levers are inactive and serve as distractors:
\begin{itemize}[leftmargin=*,noitemsep,nolistsep]
    \item \textbf{\acf{cc}}: A \textit{divergent} structure where a single first-stage lever ($L_1$) enables multiple second-stage options. Pushing $L_1$ unlocks $L_2$, $L_3$, or $L_4$, yielding three solutions: $L_1 \to \{L_2, L_3, L_4\} \to \text{Door}$.
    \item \textbf{\acf{ce}}: A \textit{convergent} structure where multiple first-stage levers funnel through a single enabler. Pushing any of $L_1$, $L_2$, or $L_3$ unlocks the same second-stage lever ($L_4$), yielding three solutions: $\{L_1, L_2, L_3\} \to L_4 \to \text{Door}$.
\end{itemize}

This design affords a clean dissociation between surface-level and structural features: across environments, lever positions, colors, and labels change, but the underlying \ac{cc} or \ac{ce} topology is preserved. Genuine structure transfer therefore requires abstracting away perceptual details to recover the invariant relational schema.

We evaluated four state-of-the-art models---GPT-5.2, Claude-4.5-Sonnet, Gemini-3-Flash, and DeepSeek-V3.2---selected to represent a diverse range of architectural paradigms and reasoning capabilities. Human behavioral data from \citet{edmonds2018human} ($N = 80$, tested under equivalent task constraints) serve as the comparative reference throughout.

\section{Experiment 1: Causal Structure Discovery}

We first investigated whether modern \ac{ai} models can discover all solutions within a \textit{single} OpenLock environment through interactive exploration, extending the causal discovery benchmark of \citet{edmonds2018human} from \ac{rl} agents to contemporary \acp{vlm} and \acp{llm}. We examined how causal structure and presentation modality jointly influence discovery trajectories.

\subsection{Experimental Design}

Following the protocol of \citet{edmonds2018human}, each model was given 30 attempts to find all solutions within a single OpenLock environment. To ensure performance stability across random environment instantiations, we tested 30 independent agents per model on each of the two causal structures (\ac{cc} and \ac{ce}). GPT-5.2, Claude-4.5-Sonnet, and Gemini-3-Flash were evaluated across all three conditions; DeepSeek-V3.2 was evaluated under the text-only condition only, as it does not support visual input in the interactive setting used here.

\paragraph{Text-only (T) Condition} Models interacted through a purely symbolic interface. Each prompt comprised: (i) high-level task objectives and operational constraints, explicitly requiring identification of all solutions; (ii) the initial state of all levers (position, color, and orientation) and door status in textual format; (iii) a sequential history of all executed actions and their outcomes; and (iv) a counter of remaining solutions. After each action, models received explicit feedback: either a null-change notification for unsuccessful attempts or a detailed state update for successful interactions (\eg, ``LOWERLEFT changes to GREY pushed''). Solution discovery was explicitly acknowledged (\eg, ``Solution found! 2 solutions remaining'').

\paragraph{Image-only (I) Condition} Models operated under a strictly visual paradigm in which all task-relevant information was conveyed through images alone. Specifically, (i) the initial environment configuration was presented solely via a representative image; (ii) lever and door state descriptions were conveyed exclusively through images; and (iii) post-action feedback consisted of dynamic visual sequences only. By removing all symbolic scaffolding, this condition isolates the contribution of pure visual input to causal discovery.

\paragraph{Text-and-Image (TI) Condition} Models received both the textual interface of the text-only condition and supplemental visual inputs. Relative to the text-only condition, this condition additionally provided (i) an image of the initial environment state alongside the textual introduction, and (ii) dynamic visual feedback reflecting environment updates after each action. This condition was designed to test whether supplemental visual information facilitates causal discovery when symbolic information is already available.

\begin{table}[b!]
    \centering
    \small
    \setlength{\tabcolsep}{3pt}
    \caption{\textbf{Experiment 1: Performance across causal structures and modalities.} Success rate (\%) and average attempt counts for \acf{cc} and \acf{ce} structures under each presentation condition. T: text-only; I: image-only; TI: text-and-image. Human data reproduced from \citet{edmonds2018human}.}
    \label{tab:exp1_full_results}
    \begin{tabular}{c c cc cc}
        \toprule
        \multirow{2}{*}{\textbf{Model}} & \multirow{2}{*}{\textbf{Condition}} & \multicolumn{2}{c}{\textbf{Success Rate (\%)}} & \multicolumn{2}{c}{\textbf{Avg.\ Attempts}} \\
        \cmidrule(lr){3-4} \cmidrule(lr){5-6}
         & & \ac{cc} & \ac{ce} & \ac{cc} & \ac{ce} \\
        \midrule
        \makecell{Human \\ {\tiny\citep{edmonds2018human}}} & --- & 65.0 & 65.0 & 19.4 & 22.0 \\
        \midrule
        \multirow{3}{*}{GPT}
         & T  & 100.0 & 66.7  & 11.8 & 19.7 \\
         & I  & 38.7  & 10.3  & 26.1 & 29.1 \\
         & TI & 66.7  & 50.0  & 22.8 & 25.4 \\
        \midrule
        \multirow{3}{*}{Claude}
         & T  & 67.7  & 86.7  & 16.9 & 14.6 \\
         & I  & 45.2  & 64.5  & 22.9 & 19.8 \\
         & TI & 86.7  & 93.3  & 17.9 & 10.2 \\
        \midrule
        \multirow{3}{*}{Gemini}
         & T  & 100.0 & 100.0 & 8.1  & 10.0 \\
         & I  & 100.0 & 100.0 & 10.0 & 12.3 \\
         & TI & 100.0 & 100.0 & 8.7  & 11.8 \\
        \midrule
        DeepSeek & T & 96.7 & 86.2 & 18.6 & 20.1 \\
        \bottomrule
    \end{tabular}
\end{table}

\subsection{Results}

\paragraph{Overall Performance}
We compared model performance in the T condition against the human baseline from \citet{edmonds2018human} (see also \cref{tab:exp1_full_results}), as this condition isolates logical inference from visual processing and thus provides the most direct comparison of causal reasoning capacity. Human participants achieved a $65.0\%$ success rate across both \ac{cc} and \ac{ce} structures, requiring an average of $20.66$ attempts ($SD=9.09$, $N=80$) to identify all three solutions. Gemini-3-Flash outperformed humans in both accuracy and efficiency, achieving a $100.0\%$ success rate with significantly fewer attempts ($M=9.08$, $SD=2.76$, $N=60$; $t(138)=9.54$, $p<.001$). GPT-5.2 and Claude-4.5-Sonnet also exceeded the human baseline in efficiency: GPT-5.2 averaged $15.77$ attempts ($SD=7.95$, $N=60$; $t(138)=3.33$, $p=.001$) with success rates between $66.7\%$ and $100.0\%$ depending on structure, and Claude-4.5-Sonnet averaged $15.74$ attempts ($SD=9.53$, $N=61$; $t(139)=3.12$, $p=.002$). DeepSeek-V3.2, while achieving high success rates ($96.7\%$ for \ac{cc}; $86.2\%$ for \ac{ce}), showed no statistically significant difference from humans in attempt count ($M=19.35$, $SD=6.63$, $N=60$; $t(138)=0.95$, $p=.346$).

\paragraph{Causal Structure Asymmetry}
Humans showed consistent success rates ($65\%$) across both structures, with no significant difference in attempt counts between \ac{cc} ($M=19.4$, $SD=9.66$, $N=40$) and \ac{ce} ($M=22.0$, $SD=8.40$, $N=40$; $t(78)=-1.27$, $p=.207$). In contrast, all models exhibited systematic asymmetries between structures. GPT-5.2 showed a strong \ac{cc} advantage in the T condition: $100.0\%$ success with $11.8$ attempts ($SD=3.38$, $N=30$) on \ac{cc}, versus $66.7\%$ success with $19.7$ attempts on \ac{ce} ($SD=9.23$, $N=30$; $t(58)=-4.38$, $p<.001$). Gemini-3-Flash similarly performed better on \ac{cc} across all conditions: despite maintaining $100\%$ success throughout, attempt counts were consistently lower for \ac{cc} ($M=8.13$, $SD=3.44$, $N=31$) than \ac{ce} ($M=10.03$, $SD=2.63$, $N=30$; $t(59)=-2.81$, $p=.0067$). Claude-4.5-Sonnet showed the opposite pattern, achieving a higher success rate on \ac{ce} ($86.7\%$) than \ac{cc} ($67.7\%$), though the difference in attempt counts was not significant ($M=14.6$, $SD=8.67$, $N=30$ for \ac{ce} \vs $M=16.9$, $SD=10.31$, $N=31$ for \ac{cc}; $t(59)=0.94$, $p=.350$). These divergent asymmetry patterns---with GPT and Gemini favoring \ac{cc} while Claude favors \ac{ce}---were absent in human learners.

\paragraph{Impact of Modality on Causal Discovery}
Adding visual information degraded performance for most models. For GPT-5.2, the TI condition required significantly more attempts than the T condition ($M=24.10$, $SD=6.95$, $N=60$ \vs $M=15.77$, $SD=7.95$, $N=60$; $t(118)=-6.11$, $p<.001$). Gemini-3-Flash showed a smaller but significant efficiency drop from T to TI ($M=9.08$, $SD=2.76$, $N=60$ \vs $M=10.41$, $SD=3.54$, $N=70$; $t(128)=-2.36$, $p=.020$), though it maintained $100\%$ success across all three conditions. Claude-4.5-Sonnet was the exception, showing no significant difference between T and TI ($M=15.74$, $SD=9.53$, $N=61$ \vs $M=14.05$, $SD=8.12$, $N=60$; $t(119)=1.05$, $p=.297$). Removing symbolic scaffolding entirely in the I condition led to further significant performance declines for GPT-5.2 (I: $M=27.55$, $SD=5.01$, $N=60$ \vs TI: $M=24.10$, $SD=6.95$, $N=60$; $t(118)=-3.12$, $p=.002$) and Claude-4.5-Sonnet (I: $M=21.37$, $SD=9.03$, $N=62$ \vs TI: $M=14.05$, $SD=8.12$, $N=60$; $t(120)=-4.70$, $p<.001$). Taken together, these results indicate that models rely primarily on symbolic text for causal reasoning, with visual input acting as a distractor rather than a facilitative cue.

\begin{figure}[t!]
    \centering
    \includegraphics[width=\linewidth]{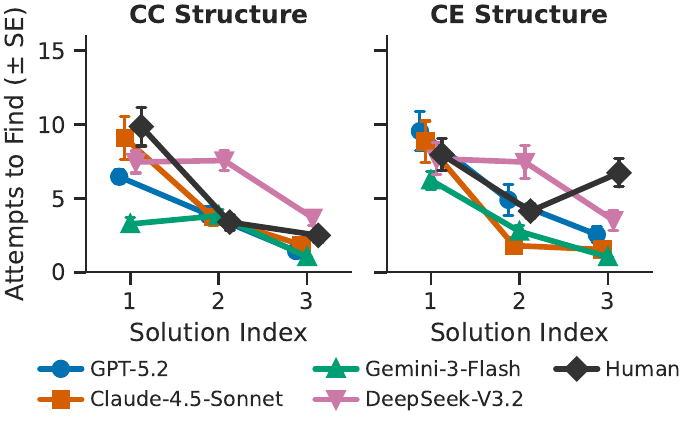}
    \caption{\textbf{Sequential discovery efficiency across causal structures.} Marginal discovery cost (attempts required to find each successive solution) within a single environment, shown separately for \ac{cc} (left) and \ac{ce} (right) structures. \ac{ai} models are evaluated in the T condition. Humans and Claude-4.5-Sonnet exhibit sharp non-linear acceleration after the first solution, whereas GPT-5.2, Gemini-3-Flash, and DeepSeek-V3.2 show more gradual improvement. Error bars denote standard error.}
    \label{fig:exp1}
\end{figure}

\paragraph{Sequential Discovery Patterns}
To characterize within-environment learning dynamics, we analyzed the marginal discovery cost---the number of attempts required to find each successive solution (see also \cref{fig:exp1}). Human learners exhibited \textbf{non-linear acceleration}: discovery cost dropped sharply from the first solution ($M=7.37$) to the second ($M=3.65$; $t(51)=4.31$, $p<.001$). Claude-4.5-Sonnet (T condition) closely mirrored this pattern, with discovery cost falling from $M=7.19$ to $M=2.62$ ($t(46)=4.80$, $p<.001$). GPT-5.2 and Gemini-3-Flash, by contrast, showed only \textbf{gradual, incremental improvement}: for example, Gemini in the T condition decreased from $M_{L1}=4.75$ to $M_{L2}=3.30$ ($t(59)=2.28$, $p=.026$), a substantially smaller reduction in magnitude than that observed in humans or Claude. These differences in learning dynamics were consistent across both \ac{cc} and \ac{ce} structures.

\section{Experiment 2: Causal Structure Transfer}

Having established baseline patterns in causal structure discovery, we investigated whether providing models with complete solutions from a structurally similar environment would facilitate discovery in a new environment---testing models' capacity for structure transfer.

\subsection{Experimental Design}

We modified the prompts for all models to include explicit textual descriptions of all three solutions from a previously completed environment with the same underlying causal structure (\ac{cc} or \ac{ce}). Each solution description specified the exact action sequence—first lever pushed, second lever pushed, and door-opening attempt (\eg, ``Solution 1: Push LOWERLEFT lever, then push UPPERLEFT lever, then try door''). The previous environment had a different spatial configuration of levers and potentially different color assignments, requiring models to abstract the structural principle beyond specific positions or visual attributes. We tested 30 agents per model per structure (240 in total: 4 models × 2 structures × 30 agents).

\begin{figure}[t!]
    \centering
    \includegraphics[width=.9\linewidth]{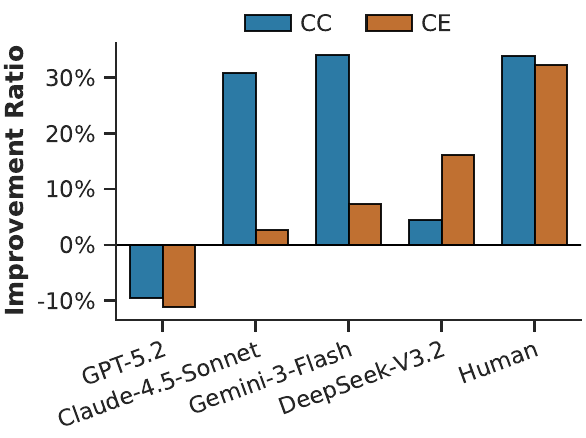}
    \caption{\textbf{Efficiency gain from causal structure transfer.} Improvement ratio in average attempt counts between Experiment 1 (Baseline) and Experiment 2 (Transfer), calculated as $(Attempts_{\text{base}} - Attempts_{\text{trans}})/Attempts_{\text{base}}$, shown separately for \ac{cc} and \ac{ce} structures. Positive values indicate positive transfer (fewer attempts required in the transfer condition).}
    \label{fig:exp2:ratio}
\end{figure}

\begin{table}[b!]
    \centering
    \small
    \setlength{\tabcolsep}{3pt}
    \caption{\textbf{Experiment 2: Overall transfer effects.} Average attempt counts in the Baseline (Experiment 1, T condition) and Transfer (Experiment 2) conditions, with improvement ratio calculated as $(Attempts_{\text{base}} - Attempts_{\text{trans}})/Attempts_{\text{base}}$. Positive improvement ratios indicate positive transfer. Human data reproduced from \citet{edmonds2018human}. $^{**}p<.01$; $^{***}p<.001$; unmarked: $p>.05$.}
    \label{tab:exp2_results}
    \resizebox{\linewidth}{!}{%
        \begin{tabular}{lccc}
            \toprule
            \textbf{Model} & \textbf{Baseline} $M$ $(SD)$ & \textbf{Transfer} $M$ $(SD)$ & \textbf{Improv.\ (\%)} \\
            \midrule
            Human {\tiny\citep{edmonds2018human}} & 20.66\ (9.09) & 13.85\ (10.00) & $+33.0^{***}$ \\
            GPT-5.2           & 15.77\ (7.95) & 17.43\ (8.61)  & $-10.5$        \\
            Claude-4.5-Sonnet & 15.74\ (9.53) & 12.92\ (8.47)  & $+17.9$        \\
            Gemini-3-Flash    &  9.08\ (2.76) &  7.33\ (3.60)  & $+19.3^{**}$   \\
            DeepSeek-V3.2     & 19.35\ (6.63) & 17.32\ (5.89)  & $+10.5$        \\
            \bottomrule
        \end{tabular}%
    }%
\end{table}

\subsection{Results}

\paragraph{Overall Transfer Effects}
The improvement ratios relative to Experiment 1 baselines are shown in \cref{fig:exp2:ratio} and summarized in \cref{tab:exp2_results}. Human participants demonstrated robust structural transfer, significantly reducing average attempts from baseline ($M=20.66$, $SD=9.09$) to transfer ($M=13.85$, $SD=10.00$; $t(158)=4.51$, $p<.001$, $d=0.71$).

\begin{figure*}[t!]
    \centering
    \includegraphics[width=\linewidth]{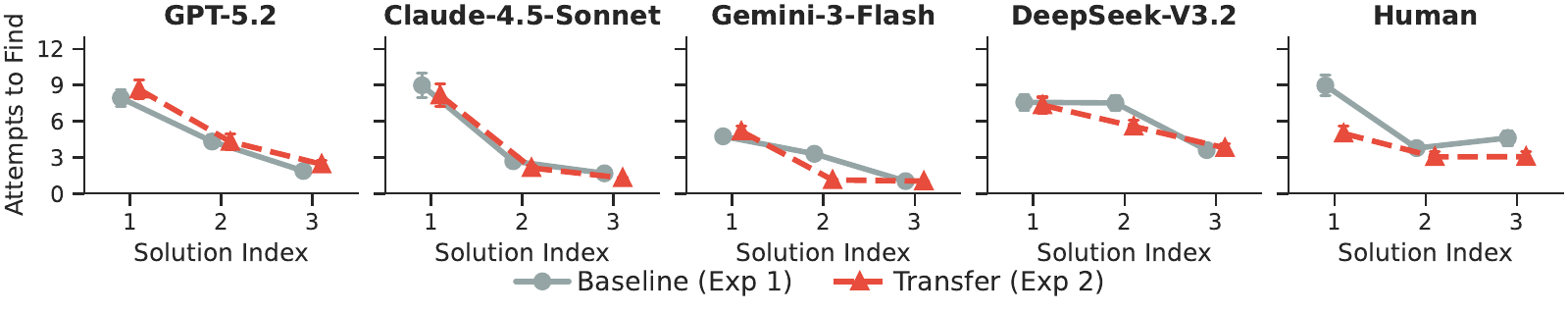}
    \caption{\textbf{Impact of causal structure transfer on sequential discovery dynamics.} Marginal discovery cost (attempts required to find each successive solution) for Baseline (Experiment 1, gray) and Transfer (Experiment 2, red) conditions, shown for each model and for human participants. Error bars denote standard error. Humans exhibit immediate transfer, with significantly lower first-solution cost under transfer. All models exhibit delayed transfer, with performance gains emerging only at the second solution (if at all), indicating that models require initial environmental grounding before leveraging prior structural knowledge.}
    \label{fig:exp2:discovery_trajectory}
\end{figure*}

In contrast, models exhibited limited transfer capabilities. Gemini-3-Flash was the only model to achieve statistically significant overall transfer, reducing average attempts from $M_{\text{base}}=9.08$ ($SD=2.76$) to $M_{\text{trans}}=7.33$ ($SD=3.60$; $t(118)=2.99$, $p=.003$, $d=0.55$). Claude-4.5-Sonnet ($M_{\text{base}}=15.74 \to M_{\text{trans}}=12.92$; $t(119)=1.71$, $p=.090$) and DeepSeek-V3.2 ($M_{\text{base}}=19.35 \to M_{\text{trans}}=17.32$; $t(118)=1.72$, $p=.089$) showed numerical trends toward improvement that did not reach statistical significance. GPT-5.2 showed no positive transfer; its attempt count numerically increased in the transfer condition ($M_{\text{base}}=15.77 \to M_{\text{trans}}=17.43$; $t(118)=-1.12$, $p=.266$), though this decline was not statistically reliable.

\paragraph{Delayed Transfer Effects in Sequential Discovery}
To characterize \textit{when} during the search process transfer effects emerge, we analyzed the marginal discovery cost for each successive solution (see also \cref{fig:exp2:discovery_trajectory}). Human participants showed immediate transfer: attempts to find the \textit{first} solution were significantly reduced from baseline ($M=8.97$, $SD=7.24$) to transfer ($M=4.99$, $SD=5.36$; $t(145)=3.82$, $p<.001$, $d=0.63$). In stark contrast, none of the four models showed a significant reduction in first-solution discovery cost, indicating that prior structural knowledge did not guide initial exploration of the new environment.

Transfer effects in models emerged only at the \textit{second} solution. Gemini-3-Flash showed a dramatic reduction in second-solution discovery cost from baseline ($M_{\text{base}}=3.30$, $SD=2.41$) to transfer ($M_{\text{trans}}=1.13$, $SD=0.39$; $t(118)=6.88$, $p<.001$, $d=1.26$). DeepSeek-V3.2 similarly showed significant acceleration ($M_{\text{base}}=7.52$, $SD=4.73$ to $M_{\text{trans}}=5.57$, $SD=3.66$; $t(110)=2.44$, $p=.016$, $d=0.46$). GPT-5.2 and Claude-4.5-Sonnet did not show statistically significant improvements at the second solution. Thus, models that did benefit from prior structural knowledge did so only after independently discovering an initial solution in the new environment, in direct contrast to humans who leveraged structural knowledge from the very first attempt.

\section{Discussion}

\subsection{Immediate \vs Delayed Transfer}

Humans immediately applied prior structural knowledge to first-solution discovery in new environments, whereas all \ac{ai} models showed delayed or absent transfer---requiring initial environmental grounding before efficiency gains emerged (\cref{fig:exp2:discovery_trajectory}). This contrast suggests that humans construct decontextualized causal schemas that directly guide action in novel contexts, while models must first establish mappings between surface tokens and structural roles through direct environmental interaction before latent structural knowledge becomes operative. Rather than possessing a fully portable causal schema, current \acp{llm} appear to exhibit a \textit{grounding-dependent transfer mechanism}: structural knowledge acquired from prior experience remains latent until activated by concrete situational feedback. Delayed transfer could also partly reflect in-context learning effects \citep{min2022rethinking} rather than a grounding-specific mechanism. This implies that for current \acp{llm}, the instantiation of abstract rules in novel contexts is a process remaining critically sensitive to situational grounding cues rather than an immediate byproduct of scale.

This pattern contrasts sharply with classic findings in human analogical reasoning, where successful transfer depends on recognizing structural similarity between source and target domains, independent of surface features \citep{gentner1983structure,holyoak1996mental}. While humans readily form what \citet{gick1983schema} termed ``problem schemas''---abstract relational representations that transfer across perceptually distinct instantiations---our results suggest that current \ac{ai} systems remain bound to context-specific instantiations, requiring direct experience with a new environment before prior structural knowledge can be exploited. This is a meaningful distinction: human transfer is \textit{prospective} (structural knowledge guides initial exploration), whereas model transfer is \textit{retrospective}.


The \ac{cc}/\ac{ce} asymmetries across models further support this interpretation. GPT-5.2 consistently favored \ac{cc} structures while Claude-4.5-Sonnet showed superior performance on \ac{ce} configurations (\cref{tab:exp1_full_results}). This pattern suggests that models encode directional statistical regularities from training corpora, where diagnostic reasoning (tracing effects back to causes) and predictive reasoning (projecting causes forward to effects) may differ in distributional frequency, rather than forming abstract causal representations that transcend directional preference. As \citet{gentner1983structure} emphasized, genuine structure mapping should enable transfer regardless of relational direction; the asymmetries indicate that current models lack this flexibility, further evidencing their reliance on surface-level statistical associations rather than genuine structural abstraction.

\subsection{Insight \vs Gradual Optimization}

Humans and Claude-4.5-Sonnet exhibited nonlinear acceleration---a reduction in discovery cost after the first solution---while GPT-5.2 and Gemini-3-Flash showed only gradual, incremental improvement (\cref{fig:exp1}). This abrupt efficiency gain in humans is consistent with \textit{representational change} \citep{ohlsson1992information}: discovering the first solution reveals the underlying causal structure, enabling the sudden elimination of entire classes of incorrect hypotheses and a qualitative reorganization of search strategy. By contrast, the smooth improvement curves of GPT-5.2 and Gemini suggest a process of iterative statistical refinement---narrowing a broad probability distribution over possible solutions rather than a discrete restructuring of the problem representation.

The in-context learning observed here thus functions more like iterative adjustment to prompt history than the discrete logical updates characteristic of human insight. Importantly, Claude-4.5-Sonnet's trajectory more closely resembles the human pattern, raising the question of whether this reflects architectural differences that better support flexible hypothesis revision, or an artifact of different search heuristics. Resolving this requires systematic investigation into how training objectives and architectural choices shape within-context learning dynamics \citep{nakkiran2020deep}.

\subsection{Multimodal Interference and the Abstraction Gap}

The addition of visual information degraded performance for most models, with the image-only condition yielding the worst results overall. This finding reveals that while humans can selectively attend to task-relevant modalities and suppress irrelevant perceptual input \citep{shams2008benefits}, current \acp{vlm} appear to lack the hierarchical control necessary to filter low-level visual features when abstract symbolic reasoning is required. In the OpenLock task, causal rules are fully determined by relational structure---lever positions, colors, and geometries are perceptually salient but causally irrelevant. Rather than providing useful abstraction support, these visual features appear to compete for processing capacity, obscuring the underlying symbolic structure.

This failure reveals a broader limitation of current multimodal architectures: an inability to distinguish between high-level causal invariants and low-level perceptual variance. Human causal reasoning relies on what might be termed \textit{abstraction through subtraction}---the capacity to ignore specific visual appearances in order to isolate invariant relational rules. Current \acp{vlm}, by contrast, appear to perform undifferentiated fusion of visual and linguistic inputs, forcing the reasoning process to integrate perceptual noise into causal hypotheses. The contrast between Claude-4.5-Sonnet---which showed no significant performance cost from added visual input---and GPT-5.2 and Gemini-3-Flash---which showed degradation---suggests that models differ in their ability to de-weight irrelevant visual information, though none achieved positive visual facilitation. Effective multimodal causal reasoning may therefore require architectures in which symbolic abstraction governs primary reasoning, with visual input serving a secondary verification role.

\subsection{Implications and Future Directions}

Together, these three findings---delayed transfer, absence of insight-like restructuring, and multimodal interference---converge on a shared conclusion: large-scale statistical learning over text and image corpora does not, by itself, produce the flexible, decontextualized causal representations that underpin human structural abstraction. Current \acp{llm} and \acp{vlm} excel at local causal search within a single context, but fail to apply structural knowledge prospectively when entering new environments. This gap is not a matter of scale or data, but appears qualitative, reflecting a fundamental difference in how humans and current \ac{ai} systems represent and deploy abstract relational structure.

These findings point toward several concrete directions for future work. First, structure-mapping curricula---training regimes that explicitly reward transfer across perceptually distinct instantiations of the same relational schema---may help bridge the prospective/retrospective transfer gap identified here. Second, the modality interference results suggest that multimodal architectures may benefit from more explicit mechanisms for cross-modal attention control, allowing visual input to inform rather than distort symbolic reasoning. Third, the divergent learning dynamics across models (Claude \vs GPT and Gemini) suggest that architectural and training choices meaningfully shape within-context learning, warranting systematic study. More broadly, the OpenLock paradigm offers a reusable benchmark for probing structural generalization in interactive settings---one that dissociates genuine abstraction from context-bound pattern matching in a way that static benchmarks cannot.

\section{Conclusion}

By comparing human and \ac{ai} performance in causal structure discovery and transfer using the OpenLock paradigm, we identified three fundamental differences in how current \acp{llm} and \acp{vlm} differ from humans in abstract causal reasoning. Humans demonstrate immediate transfer of structural knowledge, show rapid nonlinear acceleration consistent with sudden representational insight, and leverage multimodal information more selectively than current \ac{ai} systems. Despite strong performance on static reasoning benchmarks, none of these capacities were reliably present in state-of-the-art models. Our findings suggest that large-scale statistical learning does not inherently produce the flexible, decontextualized causal schemas that characterize human structural abstraction---the gap between humans and current \ac{ai} is not merely quantitative, but qualitative. This work points toward concrete development paths---including structure-mapping curricula and architectures with explicit cross-modal attention control---for building \ac{ai} systems capable of human-like causal abstraction.

\paragraph{Acknowledgment}
This work is supported in part by the National Science and Technology Major Project (2022ZD0114900), National Natural Science Foundation of China (62376009), the PKU-BingJi Joint Laboratory for Artificial Intelligence, the Wuhan Major Scientific and Technological Special Program (2025060902020304), the Hubei Embodied Intelligence Foundation Model Research and Development Program, and the National Comprehensive Experimental Base for Governance of Intelligent Society, Wuhan East Lake High-Tech Development Zone.

\printbibliography

@inproceedings{brown2020language,
  title={Language models are few-shot learners},
  author={Brown, Tom and Mann, Benjamin and Ryder, Nick and Subbiah, Melanie and Kaplan, Jared D and Dhariwal, Prafulla and Neelakantan, Arvind and Shyam, Pranav and Sastry, Girish and Askell, Amanda and others},
  booktitle=NIPS,
  year={2020}
}

@article{achiam2023gpt,
  title={GPT-4 technical report},
  author={Achiam, Josh and Adler, Steven and Agarwal, Sandhini and Ahmad, Lama and Akkaya, Ilge and Aleman, Florencia Leoni and Almeida, Diogo and Altenschmidt, Janko and Altman, Sam and Anadkat, Shyamal and others},
  journal={arXiv preprint arXiv:2303.08774},
  year={2023}
}

@book{holyoak1996mental,
  title={Mental leaps: Analogy in creative thought},
  author={Holyoak, Keith J and Thagard, Paul},
  year={1996},
  publisher={MIT press}
}

@article{holyoak2010analogical,
  title={Analogical and category-based inference: a theoretical integration with Bayesian causal models.},
  author={Holyoak, Keith J and Lee, Hee Seung and Lu, Hongjing},
  journal={Journal of Experimental Psychology: General},
  volume={139},
  number={4},
  pages={702},
  year={2010},
  publisher={American Psychological Association}
}

@article{kiciman2023causal,
  title={Causal reasoning and large language models: Opening a new frontier for causality},
  author={Kiciman, Emre and Ness, Robert and Sharma, Amit and Tan, Chenhao},
  journal={Transactions on Machine Learning Research},
  year={2023}
}

@inproceedings{jin2024can,
  title={Can Large Language Models Infer Causation from Correlation?},
  author={Jin, Zhijing and Liu, Jiarui and Lyu, Zhiheng and Poff, Spencer and Sachan, Mrinmaya and Mihalcea, Rada and Diab, Mona T and Sch{\"o}lkopf, Bernhard},
  booktitle=ICLR,
  year={2024}
}

@inproceedings{jin2023cladder,
  title={Cladder: Assessing causal reasoning in language models},
  author={Jin, Zhijing and Chen, Yuen and Leeb, Felix and Gresele, Luigi and Kamal, Ojasv and Lyu, Zhiheng and Blin, Kevin and Gonzalez Adauto, Fernando and Kleiman-Weiner, Max and Sachan, Mrinmaya and others},
  booktitle=NIPS,
  year={2023}
}

@inproceedings{edmonds2019decomposing,
  title={Decomposing human causal learning: Bottom-up associative learning and top-down schema reasoning},
  author={Edmonds, Mark and Qi, Siyuan and Zhu, Yixin and Kubricht, James and Zhu, Song-Chun and Lu, Hongjing},
  booktitle=CogSci,
  year={2019}
}

@article{lake2017building,
  title={Building machines that learn and think like people},
  author={Lake, Brenden M and Ullman, Tomer D and Tenenbaum, Joshua B and Gershman, Samuel J},
  journal={Behavioral and Brain Sciences},
  volume={40},
  year={2017},
  publisher={Cambridge University Press}
}

@article{openai2023gpt4v,
  title={GPT-4V(ision) system card},
  author={{OpenAI}},
  journal={OpenAI Technical Report},
  year={2023}
}

@article{anthropic2024claude,
  title={Claude 3 model card},
  author={{Anthropic}},
  journal={Anthropic Technical Report},
  year={2024}
}

@article{team2023gemini,
  title={Gemini: A family of highly capable multimodal models},
  author={{Gemini Team}},
  journal={arXiv preprint arXiv:2312.11805},
  year={2023}
}

@inproceedings{edmonds2018human,
  title={Human Casual Transfer: Challenges for Deep Reinforcement Learning},
  author={Edmonds, Mark and Kubricht, James and Summers, Colin and Zhu, Yixin and Rothrock, Brandon and Zhu, Song-Chun and Lu, Hongjing},
  booktitle=CogSci,
  year={2018}
}

@inproceedings{edmonds2020theory,
  title={Theory-based causal transfer: Integrating instance-level induction and abstract-level structure learning},
  author={Edmonds, Mark and Ma, Xiaojian and Qi, Siyuan and Zhu, Yixin and Lu, Hongjing and Zhu, Song-Chun},
  booktitle=AAAI,
  year={2020}
}

@article{gick1983schema,
  title={Schema induction and analogical transfer},
  author={Gick, Mary L and Holyoak, Keith J},
  journal={Cognitive Psychology},
  volume={15},
  number={1},
  pages={1--38},
  year={1983},
  publisher={Elsevier}
}

@article{gentner1983structure,
  title={Structure-mapping: A theoretical framework for analogy},
  author={Gentner, Dedre},
  journal={Cognitive Science},
  volume={7},
  number={2},
  pages={155--170},
  year={1983},
  publisher={Elsevier}
}

@article{shams2008benefits,
  title={Benefits of multisensory learning},
  author={Shams, Ladan and Seitz, Aaron R},
  journal={Trends in Cognitive Sciences},
  volume={12},
  number={11},
  pages={411--417},
  year={2008},
  publisher={Elsevier}
}

@inproceedings{nakkiran2020deep,
  title={The Deep Bootstrap Framework: Good Online Learners are Good Offline Generalizers},
  author={Nakkiran, Preetum and Neyshabur, Behnam and Sedghi, Hanie},
  booktitle=ICLR,
  year={2020}
}

@article{ohlsson1992information,
  title={Information-processing explanations of insight and related phenomena},
  author={Ohlsson, Stellan},
  journal={Advances in the Psychology of Thinking},
  pages={1--44},
  year={1992},
  publisher={Harvester Wheatsheaf}
}

@article{holyoak2011causal,
  title={Causal learning and inference as a rational process: The new synthesis},
  author={Holyoak, Keith J and Cheng, Patricia W},
  journal={Annual Review of Psychology},
  volume={62},
  number={1},
  pages={135--163},
  year={2011},
  publisher={Annual Reviews}
}

@article{xu2025towards,
  title={Towards large reasoning models: A survey of reinforced reasoning with large language models},
  author={Xu, Fengli and Hao, Qianyue and Zong, Zefang and Wang, Jingwei and Zhang, Yunke and Wang, Jingyi and Lan, Xiaochong and Gong, Jiahui and Ouyang, Tianjian and Meng, Fanjin and others},
  journal={arXiv preprint arXiv:2501.09686},
  year={2025}
}

@article{griffiths2005structure,
  title={Structure and strength in causal induction},
  author={Griffiths, Thomas L and Tenenbaum, Joshua B},
  journal={Cognitive Psychology},
  volume={51},
  number={4},
  pages={334--384},
  year={2005},
  publisher={Elsevier}
}

@article{griffiths2009theory,
  title={Theory-based causal induction.},
  author={Griffiths, Thomas L and Tenenbaum, Joshua B},
  journal={Psychological Review},
  volume={116},
  number={4},
  pages={661},
  year={2009},
  publisher={American Psychological Association}
}

@article{lu2008bayesian,
  title={Bayesian generic priors for causal learning.},
  author={Lu, Hongjing and Yuille, Alan L and Liljeholm, Mimi and Cheng, Patricia W and Holyoak, Keith J},
  journal={Psychological Review},
  volume={115},
  number={4},
  pages={955},
  year={2008},
  publisher={American Psychological Association}
}

@article{liu2025deepseek,
  title={Deepseek-v3. 2: Pushing the frontier of open large language models},
  author={Liu, Aixin and Mei, Aoxue and Lin, Bangcai and Xue, Bing and Wang, Bingxuan and Xu, Bingzheng and Wu, Bochao and Zhang, Bowei and Lin, Chaofan and Dong, Chen and others},
  journal={arXiv preprint arXiv:2512.02556},
  year={2025}
}

@article{zhang2026proposing,
  title={Proposing and solving olympiad geometry with guided tree search},
  author={Zhang, Chi and Song, Jiajun and Li, Siyu and Liang, Yitao and Ma, Yuxi and Wang, Wei and Zhu, Yixin and Zhu, Song-Chun},
  journal={Nature Machine Intelligence},
  volume={8},
  pages={84-95},
  year={2026}
}

@inproceedings{min2022rethinking,
  title={Rethinking the role of demonstrations: What makes in-context learning work?},
  author={Min, Sewon and Lyu, Xinxi and Holtzman, Ari and Artetxe, Mikel and Lewis, Mike and Hajishirzi, Hannaneh and Zettlemoyer, Luke},
  booktitle={EMNLP},
  year={2022}
}

@string {NIPS = "{Proceedings of Advances in Neural Information Processing Systems (NeurIPS)}"}

@string {ICLR = "{Proceedings of International Conference on Learning Representations (ICLR)}"}

@string {AAAI = "{Proceedings of AAAI Conference on Artificial Intelligence (AAAI)}"}

@string {EMNLP = "{Annual Conference on Empirical Methods in Natural Language Processing (EMNLP)}"}

@string {CogSci = "{Annual Meeting of the Cognitive Science Society (CogSci)}"}

@string {CHI = "{ACM Conference on Human Factors in Computing Systems (CHI)}"}

\end{document}